\documentclass{article}

\usepackage{microtype}
\usepackage{graphicx}
\usepackage{subfigure}
\usepackage{dsfont}

\usepackage{booktabs} 
\usepackage{url}
\newcommand{\indep}{\perp \!\!\!\! \perp}

\usepackage{color}
\usepackage{amsthm,amsmath,amssymb}

\newtheorem{assumption}{{Assumption}}
\newtheorem{theorem}{{Theorem}}
\newtheorem{corollary}{{Corollary}} 
\usepackage{hyperref}

\usepackage[accepted]{icml2021}

\icmltitlerunning{Hölder Bounds for Sensitivity Analysis in Causal Reasoning}

\begin{document}

\twocolumn[
\icmltitle{Hölder Bounds for Sensitivity Analysis in Causal Reasoning}




\begin{icmlauthorlist}
\icmlauthor{Serge Assaad}{ece}
\icmlauthor{Shuxi Zeng}{stat}
\icmlauthor{Henry Pfister}{ece}
\icmlauthor{Fan Li}{stat}
\icmlauthor{Lawrence Carin}{ece}
\end{icmlauthorlist}

\icmlaffiliation{ece}{Department of Electrical \& Computer Engineering, Duke University}
\icmlaffiliation{stat}{Department of Statistical Science, Duke University}

\icmlcorrespondingauthor{Serge Assaad}{serge.assaad@duke.edu}

\icmlkeywords{Machine Learning, ICML}

\vskip 0.3in
]



\printAffiliationsAndNotice{}  

\begin{abstract}
We examine interval estimation of the effect of a treatment $T$ on an outcome $Y$, given the existence of an unobserved confounder $U$. Using Hölder's inequality, we derive a set of bounds on the confounding bias $|E[Y|T=t]-E[Y|do(T=t)]|$ based on the degree of unmeasured confounding (\textit{i.e.}, the strength of the connection $U\rightarrow T$, and the strength of $U\rightarrow Y$). These bounds are tight either when $U\indep T$ or $U\indep Y | T$ (when there is no unobserved confounding). We focus on a special case of this bound depending on the total variation distance between the distributions $p(U)$ and $p(U|T=t)$, as well as the maximum (over all possible values of $U$) deviation of the conditional expected outcome $E[Y|U=u,T=t]$ from the average expected outcome $E[Y|T=t]$. We discuss possible calibration strategies for this bound to get interval estimates for treatment effects, and experimentally validate the bound using synthetic and semi-synthetic datasets.
\end{abstract}

\section{Introduction}

A typical assumption made in the treatment effect estimation literature is \textit{ignorability} -- \textit{i.e.}, that there are no unobserved confounders. This is a useful assumption since it (along with other assumptions) enables point-identification of treatment effects from observed data \citep{Imbens,Pearl}. Ignorability may be more plausible for datasets where we collect an exhaustive number of covariates (\textit{e.g.}, electronic health record data \citep{EHR,jensen2012mining_EHR}), but this assumption is untestable based on observed data.

Let $Y$ be the outcome of interest and $T$ be the treatment. Relaxing the ignorability assumption, we may assume the existence of an unobserved confounder $U$, and make assumptions about the strength of $U\rightarrow T$ and $U \rightarrow Y$ to get an interval estimate of the treatment effect -- this is known as \textit{sensitivity analysis} \cite{cornfield1959smoking,rosenbaum2010_sa,robins2000sensitivity}. Some proposals for sensitivity analysis proceed with additional modeling or distributional assumptions about the unmeasured confounder \citep{bross1966spurious,schlesselman1978assessing,rosenbaum1983assessing}, which induce additional untestable assumptions. To overcome this, we make the following contributions in this work:
\begin{enumerate}
\item  Making minimal assumptions about $(U,T,Y)$, we bound the confounding bias\footnote{In this work, we use ``confounding bias'' to refer to $|E(Y|t)-E(Y|do(t))|$, though previous work has used it to mean \textit{e.g.},$[E(Y|T=1)-E(Y|T=0)]-[E(Y|do(T=1))-E(Y|do(T=0))]$ \citep{zheng2021copulabased,VanderWeele2011}.} between the observational expectation $E[Y|T=t]$ and the \textit{interventional} expectation $E[Y|do(T=t)]$ \citep{Pearl} by a \textit{treatment sensitivity parameter} (quantifying the strength of the direct connection $U\rightarrow T$), and an \textit{outcome sensitivity parameter} (quantifying the strength of the direct connection $U\rightarrow Y$). These bounds are tight when either $U\indep T$ or $U\indep Y | T$ (\textit{i.e.}, when ignorability is satisfied).
\item We examine a special case of these bounds that is relatively easy to calibrate, and apply it to obtain the interval estimates of treatment effects $E[Y|do(T=t_1)]-E[Y|do(T=t_2)]$ for any two treatments $t_1,t_2$. Our results are also applicable to \textit{conditional} average treatment effect (CATE) estimation (conditioned on observed covariates $X=x$).
\item We discuss possible calibration strategies for the bound, allowing us to find reasonable sensitivity parameter values.
\end{enumerate}

\section{Methods}
\subsection{Basic setup}
\begin{figure}
    \centering
    \includegraphics[width=0.3\columnwidth]{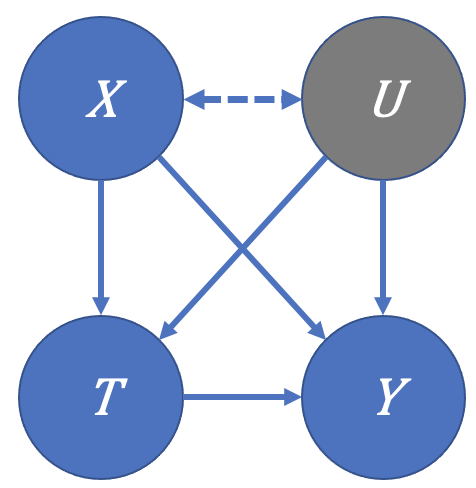}
    \caption{Assumed causal graph between unobserved confounder $U$, observed confounder $X$, treatment $T$, and outcome $Y$. Gray indicates $U$ is unobserved, and the dotted double-sided arrow indicates a possible correlation.}
    \label{fig:causal graph}
\end{figure}

Suppose we observe, for each unit $i$ (of $N$ units), a treatment $t_i\in \mathcal{T}\subset\mathbb{R}^{d_T}$, an outcome $y_i\in\mathcal{Y}\subset\mathbb{R}$, and observed covariates $x_i\in \mathcal{X}\subset\mathbb{R}^{d_X}$. Hence, our observed dataset is $\mathcal{D}=\{x_i,t_i,y_i\}_{i=1}^N$. Additionally, we assume the existence of an unobserved confounder $u_i \in \mathcal{U}\subset\mathbb{R}^{d_U}$. We use the capital letters $X,T,Y,U$ to denote the random variables for the covariates, the treatment, the outcome and the unobserved confounder, respectively. The assumed causal graph \cite{Pearl} relating these is shown in Figure \ref{fig:causal graph}.

We make the following assumptions:
\begin{assumption}[Latent ignorability] \label{assum:latent igno}
The set $\{U,X\}$ blocks all backdoor paths between $T$ and $Y$.
\end{assumption}
\begin{assumption}[Positivity]
$\forall  (t,u,x)\in \mathcal{T}\times\mathcal{U}\times\mathcal{X},~~ \textup{Pr}(T=t|U=u,X=x)>0$.
\end{assumption}

For brevity, we omit the condition $X=x$ from all conditional statements/probabilities and leave it as implicit. We also use \textit{e.g.}, $p(y|u,t)$ as shorthand for $\textup{Pr}(Y=y|U=u,T=t)$ and \textit{e.g.}, $p(Y|u,t)$ as shorthand for the density of $Y$ given $U=u,T=t$.
Under the above assumptions, we can write the interventional distribution $p(Y|do(t))$ as:
\begin{align}
    p(Y|do(t)) &= \int_\mathcal{U} p(Y|do(t),u)p(u) du \\&\overset{(*)}{=} \int_\mathcal{U} p(Y|t,u)p(u) du,\label{eq:do expansion}
\end{align}
where $(*)$ holds from Assumption \ref{assum:latent igno} \citep{Pearl,damour2019multicause}.
We can also write the observational distribution $p(Y|t)$ as:
\begin{align}
    p(Y|t) = \int_\mathcal{U} p(Y|t,u)p(u|t)du. \label{eq:cond expansion}
\end{align}

Distribution $p(Y|do(t))$ is of interest but inestimable (from observed data), and $p(Y|t)$ is estimable but uninteresting.
In general $p(Y|t) \neq p(Y|do(t))$ (the key difference is that we integrate over $p(u)$ in \eqref{eq:do expansion} vs. $p(u|t)$ in \eqref{eq:cond expansion}) -- however, there are two special cases (no unobserved confounding) where $p(Y|do(t))=p(Y|t)$:
\begin{enumerate}
    \item $U\indep T$: $p(Y|t) = \int_\mathcal{U} p(Y|t,u)p(u|t)du = \int_\mathcal{U} p(Y|t,u)p(u)du = p(Y|do(t))$.
    \item $ U \indep Y | T $: $p(Y|do(t)) = \int_\mathcal{U} p(Y|t,u)p(u)du = \int_\mathcal{U} p(Y|t)p(u)du = p(Y|t)$.
\end{enumerate}
Next, we extrapolate from the above two scenarios -- specifically, we provide a bound on $|E[Y|t]-E[Y|do(t)]|$ that vanishes when $U\indep T$ or $U\indep Y | T$.
\subsection{Hölder bounds on the confounding bias}
Here, we state our main result on bounding the confounding bias $|E[Y|t]-E[Y|do(t)]|$,
\begin{theorem} \label{thm:holder}
Assuming $$\int_\mathcal{Y,U}|y| |p(y|t)-p(y|u,t)| | p(u|t)-p(u)| du dy < \infty,$$ 
the confounding bias is bounded, for any $p,q \geq 1$ s.t. $1/p + 1/q = 1$, by:
\begin{align}
    \left|E[Y|t]-E[Y|do(t)]\right| \leq B^p_{UT}(t) \cdot B^q_{UY}(t), \label{eq:holder}
\end{align}
where $B^p_{UT}(t)$ and $B^q_{UY}(t)$ are the treatment and outcome sensitivity parameters (respectively), defined by:
\begin{align}
    B^p_{UT}(t) &\triangleq \left[\int_\mathcal{U} |p(u|t)-p(u)|^p du\right]^{1/p}; \label{eq:B_UT}\\
    B^q_{UY}(t) &\triangleq \left[\int_\mathcal{U} |E[Y|t,u]-E[Y|t]|^q du\right]^{1/q}\label{eq:B_UY}.
\end{align}
\end{theorem}
All proofs are provided in the Supplementary Material (SM, Section \ref{sec:proofs}).
Intuitively, $B^p_{UT}(t)$ quantifies the strength of the connection $U \rightarrow T$, and it is easy to see that $B^p_{UT}(t) = 0$ when $U\indep T$. Similarly, $B^q_{UY}(t)$ quantifies the strength of the connection $U \rightarrow Y$, and it is easy to see $B^q_{UY}(t)=0$ when $U\indep Y | T$. Hence, under no unobserved confounding, the bound in \eqref{eq:holder} vanishes.

\subsection{Special cases}
There are infinitely many bounds we could obtain from Theorem \ref{thm:holder}, parametrized by the choice of $p,q$ -- we focus on only one of them here, since it is relatively easy to interpret.
\begin{corollary}\label{cor:TVD}
Setting $p=1,~q=\infty$ in Theorem \ref{thm:holder}, we get:
\begin{small}
\begin{align}
&|E[Y|t]-E[Y|do(t)]| \notag\\
&\leq 2\cdot TV[p(U|t),p(U)]\cdot \underset{u\in\mathcal{U}}{\textup{sup}} |E[Y|t,u]-E[Y|t]|, \label{eq:TV bound}
\end{align}
\end{small}
where $TV$ is the total variation distance.
\end{corollary}

\textit{Remark:} Note that additional assumptions are required to guarantee that the RHS of \eqref{eq:TV bound} is finite -- it is sufficient to assume $Y$ is bounded, which guarantees that the outcome sensitivity parameter $\underset{u\in\mathcal{U}}{\textup{sup}} |E[Y|t,u]-E[Y|t]|$ is finite.

Corollary \ref{cor:TVD} bounds the confounding bias by the total-variation distance between $p(U)$ and $p(U|t)$ and the largest absolute difference between the conditional expected outcome $E[Y|t,u]$ and the average expected outcome $E[Y|t]$. We argue that this constitutes an interpretable version of Theorem \ref{thm:holder} that is relatively easy to calibrate -- we elaborate on this in Section \ref{sec:calibration}.

Finally, we can of course write the tightest bound from the class of bounds in Theorem \ref{thm:holder}:
\begin{small}
\begin{align}
    \hspace{-4mm}|E[Y|t]-E[Y|do(t)]| \leq \underset{p,q\geq 1:~ 1/p + 1/q = 1}{\textup{inf}}B^p_{UT}(t) \cdot B^q_{UY}(t).
\end{align}
\end{small}
This is an interesting optimization problem for future work.
\subsection{Treatment effect bounds}
For any two treatments $t_1,t_2\in\mathcal{T}$, we define the average treatment effect $\tau^{t_1,t_2}_{\textup{ATE}}$ and the ignorable treatment effect estimate $\tau^{t_1,t_2}_{\textup{igno}}$ as:
\begin{align}
    \tau_{\textup{ATE}}^{t_1,t_2} &\triangleq E[Y|do(t_1)]-E[Y|do(t_2)];\\
    \tau^{t_1,t_2}_{\textup{igno}} &\triangleq E[Y|t_1]-E[Y|t_2].
\end{align}
Below, we use the result in Corollary \ref{cor:TVD} to bound the average treatment effect.
\begin{corollary}\label{cor:ATE bound}
For any $t_1,t_2\in \mathcal{T}$, we have:
\begin{align}
    \tau^{t_1,t_2}_{\textup{ATE}} \in [\tau^{t_1,t_2}_{\textup{igno}} - W(t_1,t_2),\tau^{t_1,t_2}_{\textup{igno}} + W(t_1,t_2)],
\end{align} where the half-width $W(t_1,t_2)$ is defined by:
\begin{small}
\begin{align}
    W(t_1,t_2) &\triangleq B^1_{UT}(t_1)\cdot B^{\infty}_{UY}(t_1) + B^1_{UT}(t_2)\cdot B^{\infty}_{UY}(t_2).
\end{align}
\end{small}
\end{corollary}
\subsection{Calibration strategies}\label{sec:calibration}
As for any sensitivity analysis, we need to either (a) justifiably set or (b) calibrate (from observed data) the values of the sensitivity parameters: in our case, we need a strategy to calibrate the treatment sensitivity parameter $TV[p(U),p(U|t)]$ as well as the outcome sensitivity parameter $\underset{u\in\mathcal{U}}{\textup{sup}} |E[Y|t,u]-E[Y|t]|$.
\subsubsection{Calibration for ATEs}\label{sec:ate calibration}
\paragraph{Outcome sensitivity parameter}
In order to set the outcome sensitivity parameter, with the additional assumption that $Y>0$, we can rewrite Corollary \ref{cor:TVD} as:
\begin{small}
\begin{align}
&\left|\frac{E[Y|t]-E[Y|do(t)]}{E[Y|t]}\right| \notag\\
&\leq 2\cdot TV[p(U|t),p(U)]\cdot \underset{u\in\mathcal{U}}{\textup{sup}} \left|\frac{E[Y|t,u]-E[Y|t]}{E[Y|t]}\right|.\label{eq:percent bound}
\end{align}
\end{small}Here, the outcome sensitivity parameter is the maximum percent difference between the expected outcome for an individual/unit and the overall expected outcome -- we argue that this can be set by a subject-matter expert. The LHS also has a nice interpretation as the percent deviation of the observational expectation from the interventional expectation. Alternatively, we can compute $\underset{x\in\mathcal{X}}{\text{sup}}~|E(Y|t,x)-E(Y|t)|$ and make a calibration assumption that \begin{equation}\underset{u\in\mathcal{U}}{\text{sup}}~|E(Y|t,u)-E(Y|t)| \leq \underset{x\in\mathcal{X}}{\text{sup}}~|E(Y|t,x)-E(Y|t)|.\end{equation} This calibration assumption is untestable, but it is in the same vein as assumptions made in \citet{franks2019flexible,zheng2021copulabased,cinelli2020}.
\paragraph{Treatment sensitivity parameter} 
We can make another calibration assumption: that $TV[p(U),p(U|t)]\leq TV[p(X),p(X|t)]$.
$TV[p(X),p(X|t)]$ can be approximated from samples as:
\begin{small}
\begin{align}
    TV[p(X),p(X|t)] \approx \frac{1}{2N}\sum_{i=1}^N |1-\frac{p(t|x_i)}{p(t)}|, \label{eq:tv approx}
\end{align}
\end{small} where $p(t|x_i)$ can be estimated using a propensity model (\textit{e.g.}, logistic regression), and $p(t) \approx \frac{1}{N}\sum_{i=1}^N\mathds{1}(t_i=t)$. A derivation of \eqref{eq:tv approx} is provided in the SM (Section \ref{sec:TVD_approx}).
\subsubsection{Calibration for CATEs}
For convenience, we rewrite the bound in \eqref{eq:TV bound} conditioned on observed covariates $X=x$:
\begin{small}
\begin{align}
|E[Y|t,x]-E[Y|do(t),x]| \leq B^1_{UT|x}(t)\cdot B^{\infty}_{UY|x}(t),
\end{align}
\end{small}where 
\begin{align}
&B^1_{UT|x}(t) = 2\cdot TV[p(U|t,x),p(U|x)]; \label{eq:B_UT | x}\\
&B^{\infty}_{UY|x}(t)=\underset{u\in\mathcal{U}}{\textup{sup}} |E[Y|t,u,x]-E[Y|t,x]|. \label{eq:B_UY | x}
\end{align}
The calibration strategies discussed in Section \ref{sec:ate calibration} work for the average $E[Y|do(t)]$, but more careful treatment is required to calibrate bounds for $E[Y|do(t),x]$ for a specific covariate value $x$. For this purpose, we borrow the ideas from \citet{zheng2021copulabased,cinelli2020}.
\paragraph{Outcome sensitivity parameter}
To calibrate the outcome sensitivity parameter, we can ``hide'' the $j$-th observed confounder dimension. We can compute the maximum absolute difference between the ``complete'' expectation from the ``incomplete'' expectation (akin to equation \eqref{eq:B_UY | x}):\begin{small}\begin{align}\rho_j(x,t) \triangleq \underset{\tilde{x}^{(j)}}{\textup{sup}}~\left|E[Y|t,\tilde{x}^{(j)},x^{(-j)}]-E[Y|t,x^{(-j)}]\right|,\end{align}\end{small}where $E[Y|t,x^{(j)}]$ and $E[Y|t,\tilde{x}^{(j)},x^{(-j)}]$ can be estimated via regression. Finally, we can make a calibration assumption: \begin{align}B^{\infty}_{UY|x}(t)\leq\underset{j}{\textup{max}}~\rho_j(x,t), \label{eq:B_UY|x calibration}\end{align}where the max is taken over all dimensions $j$ of the observed covariates.

\paragraph{Treatment sensitivity parameter} For the treatment sensitivity parameter, we can approximate $TV\left[p(X^{(j)}|x^{(-j)}),p(X^{(j)}|x^{(-j)},t)\right]$ (akin to eq. \eqref{eq:B_UT | x}) via:
\begin{small}
\begin{align}
    TV\left[p(X^{(j)}|x^{(-j)}),p(X^{(j)}|x^{(-j)},t)\right] \approx \frac{1}{2}\left|1-\frac{p(t|x)}{p(t|x^{(-j)})}\right|, \label{eq:tv approx|x}
\end{align}\end{small}where $j$ represents a ``hidden'' covariate dimension and $p(t|x)$, $p(t|x^{(-j)})$ are estimated via logistic regression. A derivation of the above approximation is provided in the SM (Section \ref{sec:TVD_approx}). Finally, we make the following calibration assumption:
\begin{small}
\begin{align}
    &TV\left[p(U|x),p(U|x,t)\right] \notag\\&\leq \underset{j}{\textup{max}}~TV\left[p(X^{(j)}|x^{(-j)}),p(X^{(j)}|x^{(-j)},t)\right]. \label{eq:B_UT|x calibration}
\end{align}
\end{small}There are practical concerns with the above proposals for CATE interval calibration:
\begin{enumerate}
    \item Taking a max over all covariate dimensions $j$ (as in \eqref{eq:B_UY|x calibration} and \eqref{eq:B_UT|x calibration}) is computationally costly, particularly for high-dimensional covariates, as it requires training separate regression/propensity models for each $j$. 
    \item Computing $\rho_j(x,t)$ for a single $j$ requires maximization over $\tilde{x}^{(j)}$ while fixing $x^{(-j)}$ in the expectation $E[Y|t,X^{(j)}=\tilde{x}^{(j)}, X^{(-j)}=x^{(-j)}]$ (approximated via a regression model over all observed covariates). We can find all the unique values of $\tilde{x}^{(j)}$ in the dataset, then take the max over those unique values.
\end{enumerate}

\section{Related Work}
There is an extensive body of literature on sensitivity analysis to the ignorability assumption \citep{robins2000sensitivity,mccandless2007bayesian,VanderWeele2011,lee2011bounding}. Most proposals, similar to ours, assume a ``strength'' of $U\rightarrow T$ and $U\rightarrow Y$ (under different definitions of ``strength'') and examine the deviation of a causal estimand of interest from a ``naive'' estimate (\textit{i.e.}, one that assumes ignorability) based on the assumed strength parameters. \citet{ding2015sensitivity} provide a lower bound on the true risk ratio based on two ratio-scale sensitivity parameters (one treatment sensitivity parameter, and one outcome sensitivity parameter) -- they also provide a lower bound on the risk difference based on these same parameters. \citet{franks2019flexible} propose a framework for flexible modeling of the observed outcomes and relate the observed and unobserved potential outcome distributions via Tukey's factorization. \citet{zheng2021copulabased} use a copula parametrization to relate the interventional distribution to the observational distribution and show that, under some assumptions about the data-generating process and in the multi-cause setting, we can identify the treatment sensitivity parameter up to a ``causal equivalence class''. \citet{kallus19a} and \citet{jesson2021quantifying} use an odds ratio between the complete propensity and nominal propensity to quantify the strength of unobserved confounding, and make assumptions about its magnitude to bound treatment effect estimates. Closely related to our work are the confounding bias formulas in \citet{VanderWeele2011}, where the authors provide formulas for the difference between ``naive'' effect estimates and true estimates. While this bias is an exact difference (and not a bound), it is difficult to calibrate against observed data, since one has to make assumptions about the distributions of $U,T,Y,X$ -- in contrast this work proposes bounds on the bias (not an exact bias formula) by the product of only two scalars, each of which can be calibrated against observed data.

\section{Experiments}
\subsection{Binary/Categorical $U,Y,T$}\label{sec:binary_UT}
\begin{figure}[h]
    \centering
    \includegraphics[width=0.9\columnwidth]{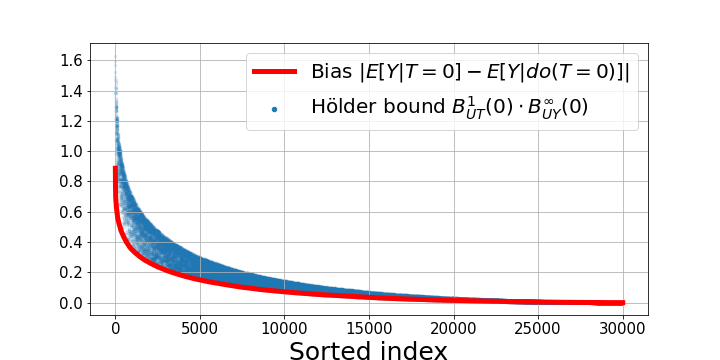}
    \vspace{-3mm}
    \caption{Confounding bias and Hölder bound, vs. the index of the sampled distribution (sorted by the confounding bias value). WLOG, we plot the bias for $T=0$.}
    \label{fig:expectation_diff_vs_TVD_sup}
\end{figure}

Let $U,Y,T$ be binary. We perform the following experiment:
\begin{itemize}
\vspace{-4mm}
    \item Draw 30,000 joint distributions $p(U,Y,T)$ from a Dirichlet($0.4\cdot\mathds{1}_8$).
    \vspace{-3mm}
    \item For each drawn $p(U,Y,T)$ and for all $t\in\{0,1\}$, compute the: $(i)$ bias $|E[Y|t]-E[Y|do(t)]|$, $(ii)$ outcome sensitivity parameter $\underset{u}{\textup{sup}} |E[Y|u,t]-E[Y|t]|$, $(iii)$ treatment sensitivity parameter $2\cdot TV[p(U),p(U|t)]$.
\end{itemize}
Figure \ref{fig:expectation_diff_vs_TVD_sup} shows the bound from Corollary \ref{cor:TVD} and the confounding bias for all sampled distributions $p(U,Y,T)$. We see that, for every bias value, we can find a joint distribution $p(U,Y,T)$ for which the bound is close to the true bias (in the binary case) -- we will more thoroughly explore bound tightness in future work.

\begin{figure}[h]
    \centering
    \includegraphics[width=0.9\columnwidth]{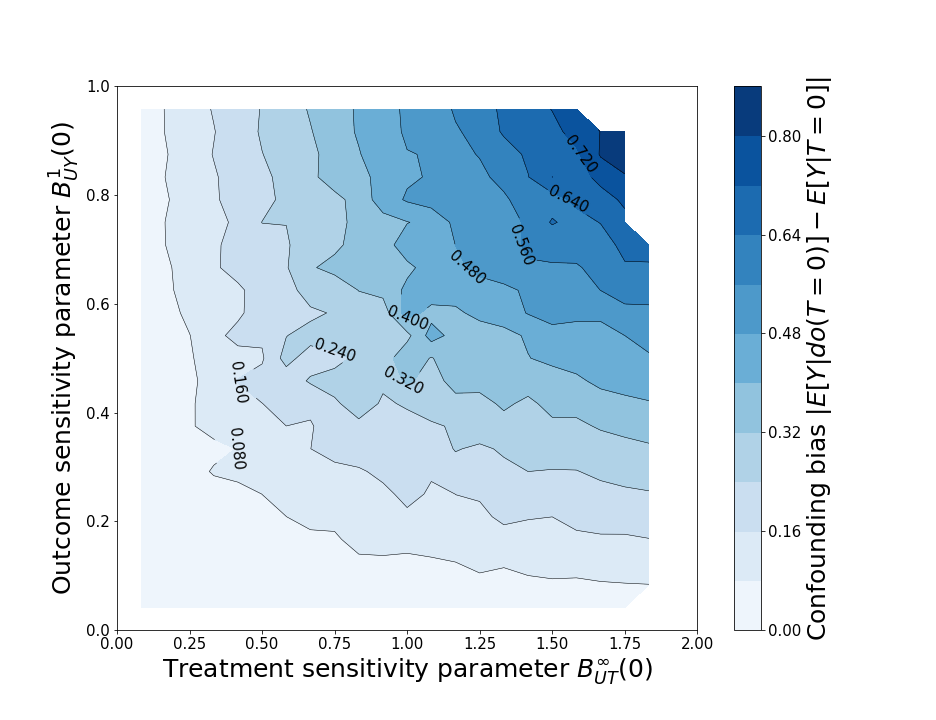}
    \vspace{-5mm}
    \caption{Confounding bias vs. treatment sensitivity parameter and outcome sensitivity parameter. WLOG, we plot the bias for $T=0$.}
    \label{fig:expectation_diff_vs_TVD_sup_heatmap}
\end{figure}

Figure \ref{fig:expectation_diff_vs_TVD_sup_heatmap} is a contour plot of confounding bias vs. treatment and outcome sensitivity parameters -- it shows that the confounding bias has an increasing trend with both sensitivity parameters, suggesting they are of equal importance in bounding the bias.
We perform the same experiment for categorical $U,T$ -- the results are shown in the SM (Section \ref{sec:categorical_UT}).
\vspace{-2mm}
\subsection{IHDP dataset}
We perform experiments on the Infant Health and Development Program (IHDP) dataset \cite{Hill}, which is semi-simulated (\textit{i.e.}, measured covariates but synthetic outcomes) and measures the effect of trained provider visits on children's test scores. There are 100 datasets within IHDP\footnote{downloaded from \href{https://www.fredjo.com}{https://www.fredjo.com}}, each with an index $d\in\{1,...,100\}$.
Similar to \cite{jesson2021quantifying}, we induce hidden confounding by hiding one of the covariates (specifically, $x_9$).
\subsubsection{ATE interval estimation}
For ATE estimation on the IHDP dataset, we first compute the naïve/ignorable ATE estimate:
\begin{small}
\begin{align}
    \hspace{-5mm} \hat{\tau}_{\textup{igno}}\triangleq \frac{1}{N_1}\sum_{i: t_i=1} y_i - \frac{1}{N_0}\sum_{i: t_i=0} y_i,
\end{align}
\end{small}where $N_1 = \sum_{i=1}^N t_i$ and $N_0 = \sum_{i=1}^N (1-t_i)$.
Next, we compute the calibrated treatment and outcome sensitivity parameters via:
\begin{align}
    &\hat{B}^1_{UT}(t) = \frac{1}{N}\sum_{i=1}^N \left|1-\frac{p(t|x_i)}{p(t)}\right|;\\
    &\hat{B}^\infty_{UY}(t) = \underset{x\in\mathcal{D}}{\text{max}}\left|\frac{1}{N_t}\sum_{i:t_i=t} y_i - E[Y|t,x]\right|,
\end{align}where $p(t)\approx\frac{1}{N}\sum_{i=1}^N\mathds{1}(t_i=t)$, $p(t|x_i)$ is estimated via logistic regression, and $E[Y|t,x]$ is estimated using a TARNet \cite{Shalit} regression model. For details on hyperparameter settings, see the SM (Section \ref{sec:hyperparams}).
Finally, from Corollary \ref{cor:ATE bound}, we compute the calibrated interval as:
\begin{align}
    \hat{I} \triangleq [\hat{\tau}_{\textup{igno}}-\hat{W}(0,1), \hat{\tau}_{\textup{igno}}+\hat{W}(0,1)],\label{eq: calibrated interval}
\end{align}where $$\hat{W}(t_1,t_2) \triangleq \hat{B}^1_{UT}(t_1)\cdot\hat{B}^\infty_{UY}(t_1) + \hat{B}^1_{UT}(t_2)\cdot\hat{B}^\infty_{UY}(t_2). $$

We can generalize the interval in \eqref{eq: calibrated interval} to scalar multiples $\alpha$ of the calibration half-width $\hat{W}$, as:
\begin{align}
        &\hat{I}(\alpha) \triangleq [\hat{\tau}_{\textup{igno}}-\alpha\hat{W}(0,1), \hat{\tau}_{\textup{igno}}+\alpha\hat{W}(0,1)] \label{eq: calibrated interval alpha}.
\end{align}
This interval becomes $\hat{I}$ when $\alpha=1$ and degenerates to the point-estimate $\hat{\tau}_{\textup{igno}}$ when $\alpha=0$.

\begin{figure}
    \centering
    \includegraphics[width=0.9\columnwidth]{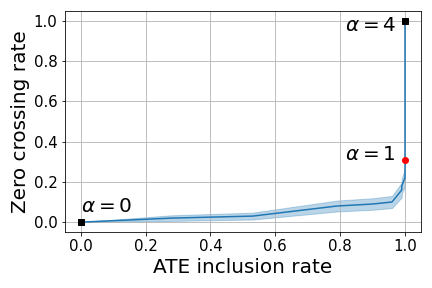}
    \caption{ATE $ZC(\alpha)$ vs. ATE $IR(\alpha)$. The points highlight the cases $\alpha=0$ (ignorable point-estimate $\hat{\tau}_{\textup{igno}}$), $\alpha=1$  (proposed interval $\hat{I}(\alpha=1)$), and $\alpha=4$ (conservative interval $\hat{I}(\alpha=4)$). The shaded area is the standard error (over 100 datasets).}
    \label{fig:ZC_vs_IR}
\end{figure}

For the IHDP dataset, we compute:
\begin{itemize}
    \item the ATE inclusion rate -- \textit{i.e.}, the percentage of datasets (out of 100 repetitions) for which the computed interval includes the true ATE. Formally, this is:
    \begin{align}
        IR(\alpha) \triangleq \frac{1}{100}\sum_{d=1}^{100} \mathds{1}[\tau_{\textup{ATE},d} \in \hat{I}_d(\alpha)],
    \end{align}where $\mathds{1}(\cdot)$ is an indicator function, $\tau_{\textup{ATE},d}$ is the true ATE for the $d$-th dataset, and $\hat{I}_d(\alpha)$ is the ATE interval for the $d$-th dataset.
    \item the ATE interval zero-crossing rate -- \textit{i.e.}, the percentage of datasets for which the computed interval contains 0:
    \begin{align}
        ZC(\alpha) \triangleq \frac{1}{100}\sum_{d=1}^{100} \mathds{1}[0 \in \hat{I}_d(\alpha)].
    \end{align}
\end{itemize}
A useful estimated ATE interval should do two things: $(i)$ include the true ATE and $(ii)$ exclude 0. Desideratum $(ii)$ is desirable because we can make a recommendation about which treatment is better on average, even under unobserved confounding. Scaling the interval (by the scalar $\alpha$) trades off the ``correctness'' of the interval (measured by $IR(\alpha)$) with its ``usefulness'' (measured by $1-ZC(\alpha)$).
We plot $ZC(\alpha)$ vs. $IR(\alpha)$ for different $\alpha$ values in Figure \ref{fig:ZC_vs_IR} -- the red point showing $\alpha=1$, \textit{i.e.}, our proposed calibrated ATE interval, which achieves a $100\%$ ATE inclusion rate, and a $31\%$ zero-crossing rate over the 100 repetitions of IHDP (with the 9th covariate hidden).
\subsubsection{CATE interval estimation}
\begin{figure}
    \centering
    \includegraphics[width=0.9\columnwidth]{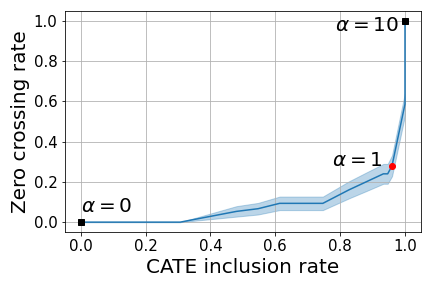}
    \caption{CATE $ZC(\alpha)$ vs. CATE $IR(\alpha)$. The points highlight the cases $\alpha=0$ (ignorable point-estimate $\hat{\tau}_{\textup{igno}}(x)$), $\alpha=1$  (proposed interval $\hat{I}_x(\alpha=1)$), and $\alpha=10$ (conservative interval $\hat{I}_x(\alpha=10)$). The shaded area is the standard error (over all samples).}
    \label{fig:ZC_vs_IR_CATE}
\end{figure}

We perform a similar experiment for CATE estimation on IHDP, this time focusing only on the first dataset $d=1$:
\begin{itemize}
    \item First, we train an ignorable model $\hat{\mu}_1(x),\hat{\mu}_0(x)$ (specifically, a TARNet) to predict the expected outcomes $E[Y|T=1,x],E[Y|T=0,x]$ (respectively) -- we define the naïve CATE estimate as $\hat{\tau}_{\textup{igno}}(x) \triangleq \hat{\mu}_1(x)-\hat{\mu}_0(x)$. We also train a logistic propensity model to approximate $p(t|x)$.
    \item Next, we compute calibrated sensitivity parameters:
    \begin{small}
    \begin{align}
        &\hat{B}^1_{UT|x}(t) = \underset{j}{\textup{max}}~\left|1-\frac{p(t|x)}{p(t|x^{(-j)})}\right|;\\
        &\hspace{-10mm}\hat{B}^\infty_{UY|x}(t) = \underset{j}{\textup{max}}\underset{\tilde{x}^{(j)}\in\mathcal{D}}{\text{max}}\left|\mu_{t}([\tilde{x}^{(j)},x^{(-j)}])-E[Y|t,x^{(-j)}]\right|.
    \end{align}
    \end{small}
    \item Finally, we compute the calibrated intervals with half-width multiplier $\alpha$:
    \begin{small}
    \begin{align}
        \hspace{-11mm}\hat{I}_x(\alpha) = [\hat{\tau}_{\textup{igno}}(x)-\alpha\hat{W}_x(0,1),\hat{\tau}_{\textup{igno}}(x)+\alpha\hat{W}_x(0,1)],
    \end{align}\end{small}where the half-width is:\begin{small}\begin{align}
        \hspace{-7mm}\hat{W}_x(t_1,t_2) \triangleq \hat{B}^1_{UT|x}(t_1)\cdot \hat{B}^\infty_{UY|x}(t_1) + \hat{B}^1_{UT|x}(t_2)\cdot \hat{B}^\infty_{UY|x}(t_2).
    \end{align}\end{small}
\end{itemize}
Similar to the previous section, we compute the following metrics:
\begin{itemize}
    \item The CATE inclusion rate -- \textit{i.e.} the percentage of samples for which the computed interval includes the true CATE \begin{small}$\tau(x)\triangleq E[Y|do(T=1),x]-E[Y|do(T=0),x]$\end{small}:
    \begin{align}
        IR(\alpha) \triangleq \frac{1}{N}\sum_{i=1}^{N} \mathds{1}[\tau(x) \in \hat{I}_x(\alpha)].
    \end{align}
    \item The CATE interval zero-crossing rate -- \textit{i.e.} the percentage of samples for which the computed interval crosses 0:
    \begin{align}
        ZC(\alpha) \triangleq \frac{1}{N}\sum_{i=1}^{N} \mathds{1}[0 \in \hat{I}_x(\alpha)].
    \end{align}
\end{itemize}
Figure \ref{fig:ZC_vs_IR_CATE} shows our results for CATE estimation on the first repetition of IHDP -- the red point shows $\alpha=1$, our proposed calibrated interval, which achieves a CATE inclusion rate of $96\%$ and a zero-crossing rate of $28\%$.

\section{Conclusions}
We have developed a bound on the confounding bias $|E[Y|do(t)]-E[Y|t]|$ based on Hölder's inequality, and used it to compute bounds on both average and conditional average treatment effects. We discussed possibilities to calibrate the sensitivity parameters in the bound, enabling practical sensitivity analysis. Finally, we performed experiments on synthetic and semi-synthetic data, showcasing empirical properties of our bound and how it can be used in practice.

This work leaves several gaps and open research directions, which we aim to explore in future work: 
\begin{enumerate}
\item What conditions on $p(U,T,X,Y)$ make our calibration assumptions (im)plausible? Empirically, we could check this by adding observed covariates $X$ to the experiment in Section \ref{sec:binary_UT}, but the question also warrants a theoretical analysis. Also, using the IHDP dataset is not an ideal ``stress test'' for our calibration assumptions, since we artificially induce hidden confounding by hiding $x_9$.
\item Can we find more computationally efficient calibration strategies for CATEs (in particular, one that doesn't require fitting many outcome models for different covariate dimensions $j$)? 
\item Can we modify the bounds to make them work with a linear-Gaussian model, as in \citet{zheng2021copulabased}? In their current form, the bounds we provide are vacuous (infinite) for the linear-Gaussian model.
\item How might we characterize the bounds' tightness? How do the bounds in this work quantitatively compare to other treatment effect bounds in the literature? 
\end{enumerate}

\bibliography{example_paper}
\bibliographystyle{icml2021}

\appendix
\section{Supplementary Material}
\subsection{Proofs}\label{sec:proofs}
\setcounter{theorem}{0}
\begin{theorem}[restated] \label{thm:holder_SM}
Assuming $$\int_\mathcal{Y,U}|y| |p(y|t)-p(y|u,t)| | p(u|t)-p(u)| du dy < \infty, $$
the confounding bias is bounded, for any $p,q \geq 1$ s.t. $1/p + 1/q = 1$, by:
\begin{align}
    |E[Y|t]-E[Y|do(t)]| \leq B^p_{UT}(t) \cdot B^q_{UY}(t),\label{eq:holder_SM}
\end{align}
where $B^p_{UT}(t)$ and $B^q_{UY}(t)$ are the treatment and outcome sensitivity parameters (respectively), defined by:
\begin{align}
    B^p_{UT}(t) &\triangleq \left[\int_\mathcal{U} |p(u|t)-p(u)|^p du\right]^{1/p}; \label{eq:B_UT_SM}\\
    B^q_{UY}(t) &\triangleq \left[\int_\mathcal{U} |E[Y|t,u]-E[Y|t]|^q du\right]^{1/q}\label{eq:B_UY_SM}.
\end{align}
\end{theorem}
\begin{proof}
We can write the confounding bias as:
\begin{align}
    &\hspace{-3mm}E[Y|t]-E[Y|do(t)] = \int_\mathcal{Y} y[p(y|t)-p(y|do(t))]dy\\
    &= \int_\mathcal{Y} y \int_\mathcal{U} p(y|u,t)[p(u|t)-p(u)]du dy.
\end{align}
Noting that $\int_\mathcal{Y} y \int_\mathcal{U} p(y|t)[p(u|t)-p(u)]du dy = \int_\mathcal{Y} y p(y|t) dy \int_\mathcal{U} [p(u|t)-p(u)] du =0$, we have:
\begin{small}
\begin{align}
    &\hspace{-5mm}E[Y|t]-E[Y|do(t)] = \int_\mathcal{Y} y \int_\mathcal{U} p(y|u,t)[p(u|t)-p(u)]du dy\notag \\&\hspace{0.8in}- \int_\mathcal{Y} y \int_\mathcal{U} p(y|t)[p(u|t)-p(u)]du dy\\ =& \int_\mathcal{Y} y \int_\mathcal{U}[p(y|u,t)-p(y|t)][p(u|t)-p(u)]du dy \\\overset{(*)}{=}& \int_\mathcal{U} [p(u|t)-p(u)]\int_\mathcal{Y} y [p(y|u,t)-p(y|t)] dy du \\
    =& \int_\mathcal{U} [p(u|t)-p(u)][E[Y|u,t]-E[Y|t]] du,
\end{align}\end{small}where $(*)$ holds by Fubini's theorem (since $\int_\mathcal{Y,U}|y| |p(y|t)-p(y|u,t)| | p(u|t)-p(u)| du dy$ is finite by assumption).
By Hölder's inequality, we have:
\begin{small}
\begin{align}
&|E[Y|t]-E[Y|do(t)]| \leq \notag\\&\hspace{-6.3mm} \left[\int_\mathcal{U} |p(u|t)-p(u)|^p du\right]^{\frac1p} \cdot \left[\int_\mathcal{U} |E[Y|t,u]-E[Y|t]|^q du\right]^{\frac1q},
\end{align}\end{small}for any $p,q\geq 1$ s.t. $1/p + 1/q = 1$.
\end{proof}
\setcounter{corollary}{1}
\begin{corollary}[restated]
For any $t_1,t_2\in \mathcal{T}$, we have:
\begin{align}
    \tau^{t_1,t_2}_{\textup{ATE}} \in [\tau^{t_1,t_2}_{\textup{igno}} - W(t_1,t_2),\tau^{t_1,t_2}_{\textup{igno}} + W(t_1,t_2)],
\end{align} where the half-width $W(t_1,t_2)$ is defined by:
\begin{small}
\begin{align}
    W(t_1,t_2) &\triangleq B^1_{UT}(t_1)\cdot B^{\infty}_{UY}(t_1) + B^1_{UT}(t_2)\cdot B^{\infty}_{UY}(t_2).
\end{align}
\end{small}
\end{corollary}
\begin{proof}
\begin{align}
\tau_{\textup{ATE}}^{t_1,t_2} &\triangleq E[Y|do(t_1)]-E[Y|do(t_2)]\\
&= \{E[Y|do(t_1)]-E[Y|t_1]\}\notag\\
&\quad+\{E[Y|t_1]-E[Y|t_2]\}\notag\\
&\quad+\{E[Y|t_2]-E[Y|do(t_2)]\}\label{eq:ATE_decomposition}\\
&\leq |E[Y|do(t_1)]-E[Y|t_1]|\notag\\
&\quad+\{E[Y|t_1]-E[Y|t_2]\}\notag\\
&\quad+|E[Y|t_2]-E[Y|do(t_2)]|\\
&\overset{(*)}\leq B^1_{UT}(t_1)\cdot B^{\infty}_{UY}(t_1)\notag\\
&\quad+\{E[Y|t_1]-E[Y|t_2]\}\notag\\
&\quad+B^1_{UT}(t_2)\cdot B^{\infty}_{UY}(t_2),
\end{align}where $(*)$ holds from Corollary \ref{cor:TVD}.
Similarly, starting from equation \eqref{eq:ATE_decomposition}:
\begin{align}
\tau_{\textup{ATE}}^{t_1,t_2} &\geq -|E[Y|do(t_1)]-E[Y|t_1]|\notag\\
&\quad+\{E[Y|t_1]-E[Y|t_2]\}\notag\\
&\quad-|E[Y|t_2]-E[Y|do(t_2)]|\\
&\overset{(*)}\geq -B^1_{UT}(t_1)\cdot B^{\infty}_{UY}(t_1)\notag\\
&\quad+\{E[Y|t_1]-E[Y|t_2]\}\notag\\
&\quad-B^1_{UT}(t_2)\cdot B^{\infty}_{UY}(t_2),
\end{align}where $(*)$ also holds from Corollary \ref{cor:TVD}.
\end{proof}
\subsection{Total Variation Distance approximations}\label{sec:TVD_approx}

We derive the approximation for the TV presented in equation \eqref{eq:tv approx}:
\begin{align}
    TV[p(X),&p(X|t)] = \frac12 \int_\mathcal{X} \left|p(x)-p(x|t)\right|dx \\&= \frac12 \int_\mathcal{X} p(x)\left|1-p(x|t)/p(x)\right|dx \\&=  \frac12 \int_\mathcal{X} p(x)\left|1-p(t|x)/p(t)\right|dx.
\end{align}
Taking MC samples from $p(X)$, we get:
\begin{align}
    TV[p(X),p(X|t)] \approx \frac{1}{2N} \sum_{i=1}^N\left|1-\frac{p(t|x_i)}{p(t)}\right|.
\end{align}
Finally, we derive the approximation for the TV presented in \eqref{eq:tv approx|x}:
\begin{align}
&TV[p(X^{(j)}|x^{(-j)}),p(X^{(j)}|x^{(-j)},t)]\\
&\hspace{-5mm}= \frac12\int_{\mathcal{X}^{(j)}} \left|p(x^{(j)}|x^{(-j)})-p(x^{(j)}|x^{(-j)},t)\right|dx^{(j)} \\&\hspace{-5mm}= \frac12 \int_{\mathcal{X}^{(j)}} p(x^{(j)}|x^{(-j)})\left|1-p(t|x^{(j)},x^{(-j)})/p(t|x^{(-j)})\right|dx^{(j)}.
\end{align} where $\mathcal{X}^{(j)} = \{x^{(j)} | x \in \mathcal{X}\}$. Taking MC samples from $p(X^{(j)}|x^{(-j)})$, we get:
\begin{align}
TV[p(X^{(j)}&|x^{(-j)}),p(X^{(j)}|x^{(-j)},t)]\notag\\
&\approx \frac{1}{2\cdot\left|\{\tilde{x}\in\mathcal{D}: \tilde{x}^{(-j)}=x^{(-j)}\}\right|}\notag\\&\cdot\sum_{\tilde{x}\in\mathcal{D}: \tilde{x}^{(-j)}=x^{(-j)}} \left|1-\frac{p(t|\tilde{x}^{(j)},x^{(-j)})}{p(t|x^{(-j)})}\right|.\label{eq:TV|x MC approx}
\end{align}
If each sample $x$ of the dataset has unique covariate values in the dimensions $x^{(-j)}$ (as is the case in IHDP, for instance), the above reduces to the one-sample approximation:
\begin{align}
&TV[p(X^{(j)}|x^{(-j)}),p(X^{(j)}|x^{(-j)},t)]\notag\\
    &\approx \frac12\left|1-\frac{p(t|x^{(j)},x^{(-j)})}{p(t|x^{(-j)})}\right| = \frac12\left|1-\frac{p(t|x)}{p(t|x^{(-j)})}\right|.
\end{align}
There are certainly more stable ways to make this approximation -- \textit{e.g.}, instead of the Monte Carlo approximation in equation \eqref{eq:TV|x MC approx}, we could take an average over all dataset samples $\tilde{x}$ such that $||\tilde{x}^{(-j)}-x^{(-j)}||\leq \epsilon$ (for some choice of norm and radius $\epsilon$), or use a kernel density estimate for $p(X^{(j)}|x^{(-j)})$.
\subsection{Additional results for categorical $U,T$}\label{sec:categorical_UT}
\begin{figure}
    \centering
    \includegraphics[width=\columnwidth]{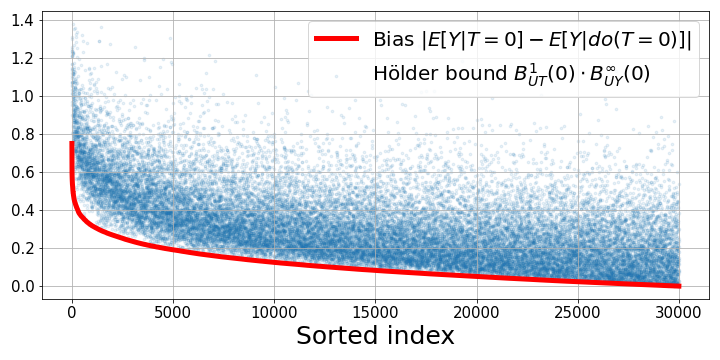}
    \caption{For $U,T$ categorical (with 4 classes) -- confounding bias and Hölder bound, vs. the index of the sampled distribution (sorted by the confounding bias value). WLOG, we plot the bias for $T=0$.}
    \label{fig:exp_bound_dim=4}
\end{figure}

\begin{figure}
    \centering
    \includegraphics[width=\columnwidth]{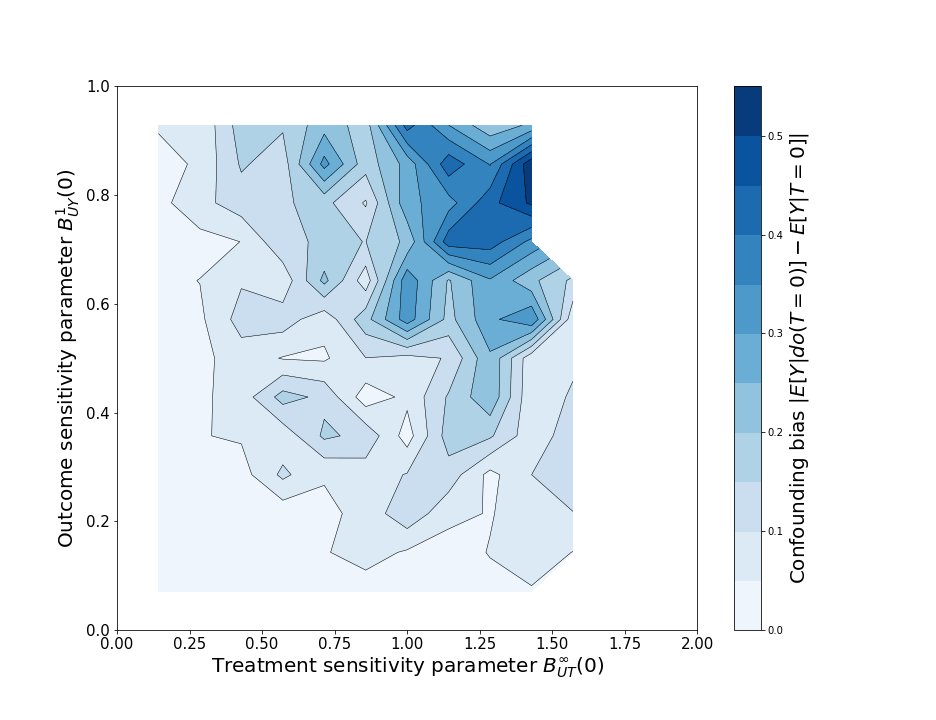}
    \caption{For $U,T$ categorical (with 4 classes) -- confounding bias vs. treatment sensitivity parameter and outcome sensitivity parameter. WLOG, we plot the bias for $T=0$.}
    \label{fig:exp_bound_contours_dim=4}
\end{figure}

We repeat the experiment in Section \ref{sec:binary_UT}, making $U$ and $T$ categorical (each with 4 categories).
The results are shown in Figures \ref{fig:exp_bound_dim=4} and \ref{fig:exp_bound_contours_dim=4}, which show similar trends as Figures \ref{fig:expectation_diff_vs_TVD_sup} and \ref{fig:expectation_diff_vs_TVD_sup_heatmap} in the main text -- \textit{i.e.}, Figure \ref{fig:exp_bound_dim=4} shows that for every bias value, we can find a joint distribution $p(U,Y,T)$ for which the bound is close to the true bias, and  Figure \ref{fig:exp_bound_contours_dim=4} shows that the confounding bias generally increases with both sensitivity parameters.
\subsection{Hyperparameters for IHDP}\label{sec:hyperparams}
The outcome model we use is an encoder $\Phi:\mathcal{X}\rightarrow\mathcal{R}$, composed with a treatment ``head'' $h_1:\mathcal{R}\rightarrow\mathcal{Y}$ and a control ``head'' $h_0:\mathcal{R}\rightarrow\mathcal{Y}$. We approximate $E(Y|T=1,X=x)$ with $h_1(\Phi(x))$ and $E(Y|T=0,X=x)$ with $h_0(\Phi(x))$. This is equivalent to TARNet \citep{Shalit}.
The hyperparameters used for the IHDP experiments are shown in Table \ref{tab:hyperparams}.
\begin{table}
    \centering
    \caption{Model hyperparameter settings for IHDP dataset.}
\begin{small}
\begin{tabular}{ll}
\toprule
     \textbf{Hyperparameter} & \textbf{Value}\\
     \midrule
     Model architecture & TARNet \citep{Shalit}\\
     Num. hidden layers in encoder $\Phi$ & 0 (i.e., $\Phi$ is a linear layer)\\
     Num. hidden layers in ``heads'' $h_1,h_0$ & 0 (linear layer)\\
     Dim. of representation space $\mathcal{R}$ & 10\\
     Activation of $\Phi$ & ReLU \\
     Batch size & $N$ (\textit{i.e.}, the whole training set)\\
    Propensity model & Logistic Regression \\
     Learning rate & 0.001 \\
     Optimizer & Adam \\
     Epochs & 5000\\
     \bottomrule
\end{tabular}
\end{small}
    \label{tab:hyperparams}
\end{table}

\end{document}